\documentclass[final]{cvpr}

\usepackage{times}
\usepackage{epsfig}
\usepackage{graphicx}
\usepackage{amsmath}
\usepackage{amssymb}
\usepackage{multirow}

\usepackage[pagebackref=true,breaklinks=true,colorlinks,bookmarks=false]{hyperref}

\begin{document}

%%%%%%%%% TITLE
\title{Relation Modeling in Spatio-Temporal Action Localization}

% \author{Yutong Feng\\
% \and

\author{
Yutong Feng$^{1,2}$
\quad Jianwen Jiang$^{2}\thanks{Corresponding authors}$ 
\quad Ziyuan Huang$^{2}$ 
\quad Zhiwu Qing$^{2}$ 
\\
\quad Xiang Wang$^{2}$ 
\quad Shiwei Zhang$^{2}$
\quad Mingqian Tang$^{2}$
\quad Yue Gao$^{1}\footnotemark[1]$ 
\\
 %Key Laboratory of Image Processing and Intelligent Control \\
$^1$THUIBCS, BNRist, School of Software, Tsinghua University\\
$^2$Alibaba Group\\

\tt\small fyt19@mails.tsinghua.edu.cn
, gaoyue@tsinghua.edu.cn\\
{\tt\small \{jianwen.jjw, pishi.hzy, qingzhiwu.qzw\}@alibaba-inc.com}\\
{\tt\small \{xiaolao.wx, zhangjin.zsw,  mingqian.tmq\}@alibaba-inc.com}
}

\maketitle

\let\thefootnote\relax\footnotetext{This work was done when Yutong Feng (Tsinghua Unviersity), Ziyuan Huang (National University of Singapore), Zhiwu Qing and Xiang Wang (Huazhong University of Science and Technology) were interns at Alibaba Group.}
%%%%%%%%% ABSTRACT

\begin{abstract}
This paper presents our solution to the AVA-Kinetics Crossover Challenge of ActivityNet workshop at CVPR 2021. Our solution utilizes multiple types of relation modeling methods for spatio-temporal action detection and adopts a training strategy to integrate multiple relation modeling in end-to-end training over the two large-scale video datasets. Learning with memory bank and finetuning for long-tailed distribution are also investigated to further improve the performance. In this paper, we detail the implementations of our solution and provide experiments results and corresponding discussions. We finally achieve 40.67 mAP on the test set of AVA-Kinetics.
% and obtain the 1st place of the AVA-Kinetics 2021 challenge. 
\end{abstract}
\section{Introduction}
Spatio-temporal action localization aims to localize atomic actions of people in videos with 3D bounding boxes, which has attract large efforts in recent years~\cite{ava,lfb,aia,acar,slowfast,tar}. Generally, there are two main factors showing fundamental influence on the performance of this task, \ie video backbones and relation modeling. The design of video networks has been widely studied \cite{slowfast, stroud2020d3d,fan2021mvit} and greatly enhance the performance of downstream tasks. Besides, pretraining such networks on large-scale networks is also demonstrated to be effective \cite{aia,acar}, \eg pretrain on Kinetics700 \cite{kinetics}. And there are also multiple ways to perform the pretraining, such as supervised pre-training~\cite{csn,slowfast,arnab2021vivit} as used in~\cite{song2019tacnet,qing2021temporal,wang2020cbr}, and unsupervised ones~\cite{huang2021self,han2020self}. For relation modeling, different approaches has been studied in the fields of computer vision \cite{wang2018non,arnab2021vivit}, social networks \cite{kipf2016semi,jiang2019dynamic} and nature language processing \cite{vaswani2017attention}. Specifically, transformed-based relation modeling has been proved for improving the spatio-temporal localization task \cite{lfb, aia,acar}.

\begin{figure*}[t]
\begin{center}
% \fbox{\rule{0pt}{1.5in} \rule{0.9\linewidth}{0pt}}
  \includegraphics[width=1\linewidth]{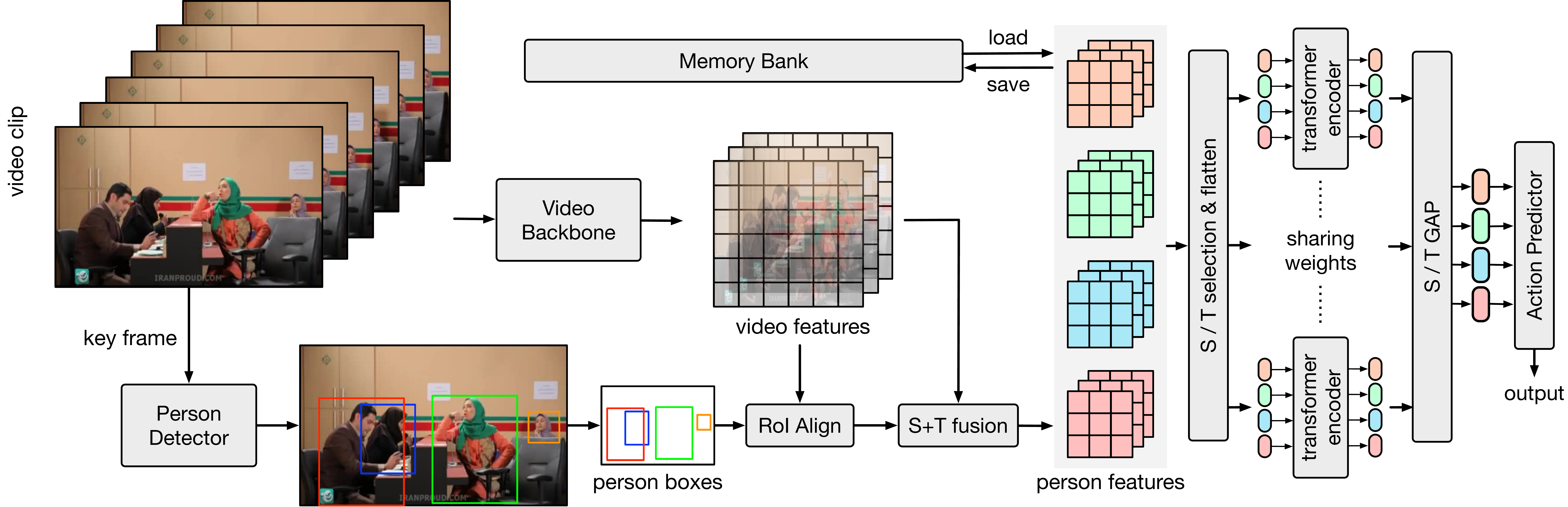}
\end{center}
   \caption{\textbf{Pipeline of our approach for spatio-temporal action localization.} `S` and `T` indicate operation along the spatial and temporal dimensions, respectively. } 
\label{fig:pipeline}
\end{figure*}

In this paper, we investigate multiple types of relation modeling methods for spatio-temporal action localization. Inspired by previous works, an off-the-shelf person detector are employed first to generate all human bounding boxes in the videos. Then we adopt a backbone model to extract visual features and build a relation module upon the feature maps of each person via roi align~\cite{maskrcnn}. After relation module, an action predictor is used to generate score for each action category. The whole pipeline of our solution is shown in ~\ref{fig:pipeline}. In following sections, we first detail the implementation of our method. Then the experimental results and corresponding discussions are provided.
\section{Method}

In this section, we present our  approach for relation modeling in spatio-temporal action localization. Firstly, we introduce our overall pipeline for this task. Then relation modeling module is presented with transformer-based architectures to capture the relations among persons in the spatial and temporal dimensions. Furthermore, we adapt memory bank for storing person features along the temporal context of video clips to model long-range relations. Different strategies of online or offline maintaining memory bank on the AVA-Kinetics Crossover are studied. Finally, we investigate learning approaches for the lone-tailed category distribution in the AVA-Kinetics dataset~\cite{ava,avakinetics}.

\subsection{Overall Pipeline}
Our designed pipeline is illustrated in Figure \ref{fig:pipeline}. Given the input video clip, the key frame of this clip is extracted and fed into a 2D person detector to generate bounding boxes of persons inside this clip. The whole video clip is sampled into frames in specified interval and encoded with an video backbone, \eg SlowFast \cite{slowfast} and CSN \cite{csn}, to output a 3D video feature map. Then the 2D person boxes are  inflated along the temporal dimension and used for extracting person features from the feature map with 3D RoI-Align \cite{maskrcnn}. The pooled person features are further fused together with the video feature map via channel-wise concatenation and convolution layers. To model the hidden relations among persons inside the same video clip for improving the effectiveness of action prediction, we fed the person features into our relation modeling module with transformer-based blocks. To specify the spatial and temporal relations, we select features along the same spatial or temporal dimension from different persons. The selected features are flatten to a sequence of tokens and fed into a transformer encoder block to model their relations via attention mechanism. Finally, the output tokens of all block in spatial or temporal dimensions are globally averaged and fed into full-connected layers to predict the action classes for each detected person.

\subsection{Person Relation Modeling}
The input person features to relation modeling module are 3D feature maps denoted as $P^i \in \mathbb{R}^{T \times H \times W}$, where $i$ is the person index. Such person features are firstly transferred into  sequential tokens as the input of transformer encoder block.
To effectively model the relations along spatial and temporal dimension while maintain low computation cost, two types of relation head, \ie S-only and T-only, are proposed to extract relations separately on each dimension. For S-only head, we generate $HW$ sequences of input tokens. The sequence corresponding to spatial position $(h, w)$ is $\{agg_{t=1}^{T}(P^i_{t, h, w}) ~|~ i \in I\}$, where $agg(\cdot)$ indicates an aggreagation function, \eg average pooling or max pooling, and $I$ is the index set of persons. Similarly, T-only head generate $T$ sequences of input tokens, where the sequence corresponding to temporal position $t$ is $\{agg_{h=1}^{H} ~_{w=1}^{W} (P^i_{t,h,w})~|~i\in I\}$.

Each sequence of tokens are then fed into a transformer encoder block for relation modeling. It is noted that transformer blocks for different spatial or temporal positions share the same weights. The output of transformer blocks are then averaged along different spatial or temporal positions into the final representation of persons.

\begin{table*}[t]
\begin{center}
\renewcommand{\arraystretch}{1.2}
\begin{tabular}{c|c|c|c|c|c|c|c|c}
\multirow{2}{*}{backbone} & \multirow{2}{*}{pretrain} & \multirow{2}{*}{input} & \multirow{2}{*}{head} & \multirow{2}{*}{+M} & \multicolumn{3}{c}{val mAP@0.5} \vline & test mAP@0.5 \\
\cline{6-9} 
 & &  &  & & AVA  & Kinetics & A+K  & A+K\\
\hline
\hline
SlowFast-152 & K700 & 8x8+32x2 & Linear & $\times$ & 35.29 & 32.88 & 36.26 & -\\
SlowFast-152 & K700 & 8x8+32x2 & S-only & $\times$ & 36.16 & 33.82 & 36.92 & -\\
SlowFast-152 & K700 & 8x8+32x2 & T-only & $\times$ & 36.34 & 33.16 & 36.88 & -\\
SlowFast-152 & K700 & 8x8+32x2 & T-only & $\surd$ & 36.88 & 33.44 & 37.35 & -\\
\hline
% SlowFast-152 & K700 & 16x4+64x1 & S-only & $\times$ & 37.29 & 33.63 & 37.64 & -\\
% SlowFast-152 & K700 & 16x4+64x1 & T-only & $\times$ & 36.82 & 34.43 & 37.48 & -\\
SlowFast-152 & K700 & 16x4+64x1 & T-only & $\surd$ & 37.68 & 35.54 & 38.12 & -\\
\hline
\hline
ir-CSN-152 & K400+IG65M & 32x2 & Linear & $\times$ & 35.38 & 32..58 & 36.20 & - \\
ir-CSN-152 & K400+IG65M & 32x2 & S-only & $\times$ & 36.07 & 32.62 & 36.58 & - \\
ir-CSN-152 & K400+IG65M & 32x2 & T-only & $\times$ & 36.29 & 32.81 & 36.58 & - \\
ir-CSN-152 & K400+IG65M & 32x2 & T-only & $\surd$ & 36.99 & 33.44 & 37.34 & - \\
\hline
ir-CSN-152 & K700 & 32x2 & Linear & $\times$ & 35.67 & 33.20 & 36.22 & - \\
ir-CSN-152 & K700 & 32x2 & S-only & $\times$ & 36.69 & 34.12 & 37.16 & - \\
ir-CSN-152 & K700 & 32x2 & T-only & $\times$ & 36.41 & 34.80 & 36.98 & - \\
ir-CSN-152 & K700 & 32x2 & T-only & $\surd$ & 37.32 & 35.01 & 37.78 & - \\
\hline
% ir-CSN-152 & K700 & 64x1 & Linear & $\times$ & 36.18 & 33.92 & 37.04 & -\\
% ir-CSN-152 & K700 & 64x1 & S-only & $\times$ & 37.28 & 34.26 & 37.92 & -\\
% ir-CSN-152 & K700 & 64x1 & T-only & $\times$ & 37.18 & 34.71 & 37.75 & -\\
ir-CSN-152 & K700 & 64x1 & T-only & $\surd$ & 37.95 & 35.26 & 38.43 & -\\
\hline
\hline
ensemble & - & - & mixed & - & \textbf{40.52} & \textbf{37.04} & \textbf{40.97} & \textbf{40.67} \\
\end{tabular}
\end{center}
\caption{\textbf{Main results on AVA-Kinetics v1.0.} All models are tested with 3 scales and horizontal flips. "+M" indicates training with memory. "A+K" indicates results on AVA-Kinetics.}
\label{tab:main}
\end{table*}

\subsection{Memory Bank}
Maintaining feature memory bank for storing and utilizing representations along long term context has been demonstrated to be effective strategies for this task \cite{lfb, aia,acar}. We also adapt the feature bank, which saves our pooled feature features and provides previously stored person features of timestamps within a long-range of current video clip. The loaded features are concatenated and fed into the relation modeling module.

Existing methods use online maintaining strategy of memory bank, which continually updates the stored features of current video during the training stage. However, for the AVA-Kinetics Crossover, such an online strategy is hard to be implemented since there is only one officially annotated clip of each video in Kinetics and the remaining clips will not be reached in the training stage. To address this issue, we design a two-stage training strategy. In the first stage, the non-annotated clips in Kinetics are not considered. We either train only on the AVA dataset or maintain an empty memory bank for the Kinetics dataset. In the second stage, we extract and store features of all clips in Kinetics, freeze the weights of backbone and finetune the relation head and classifier on both AVA and Kinetics. Besides, we also investigate the strategy that training without memory bank in the first stage, and retrain another head with memory bank in the second stage. Comparison of different strategies are shown in experiments.

\subsection{Long-tailed Learning}
There exists obvious long-tailed category distribution in the original AVA dataset, which leads to the challenge of learning those classes with less number of samples. With the join of Kinetics annotated data, the long-tailed distribution still remains a large problem for this task.

To perform a more suitable training, we consider the decoupling strategy from \cite{kang2019decoupling}. The training process is decoupled into two stages. Stage one follows the normal training strategy with randomly sampled data. While in stage two, we freeze all the models except the final classifier and train with class-balanced data sampling. Such a strategy helps to improve the performance on small classes. 

\section{Experiments}

\subsection{Experimental Settings}
\textbf{Person Detector.} We adopt GFocalV2~\cite{gflv2} as the person detector. We first train the model on the subset of person category from COCO Dataset~\cite{coco}. Then we continue training the model on the AVA-Kinetics dataset from COCO-Person pre-training.
\par
\textbf{Backbone.} We adopt CSN152~\cite{csn}, slowfast101~\cite{slowfast}, slowfast152~\cite{slowfast} as the visual feature extractors. We first train the model on the training set of Kinetics700~\cite{kinetics} dataset and then use the weight to initial the backbone part of our pipeline.
\par
\textbf{Heads.} We train with four types of heads to predict actions. The linear head simply use full-connected layers as the baseline head. For relation heads, considering the size of features from S-only head will make it changeable to maintain memory bank , we train S-only without memory bank and T-only with/without memory bank.
\par
\textbf{Training and Inference.} During the training process, we concatenate the data list of AVA and Kinetics for mixed learning. We train with SGD optimizer, initial learning rate as 1e-2, batch size as 64,  weight decay as 1e-7 and total training iterations as 30k. All the batchnorm layers are frozen in training. We do linear warmup in the first 1.5k iterations, drop the learning rate by 0.66 at iteration 13.5k, 18k, 22.5k and 27k. The input videos are resized with minimum side of 256 and maximum side of 464. Color jitter and box jitter are used for data augmentation. During the inference process, we test with three scales \{256, 288 and 320\} and horizontal flips.
\begin{table*}[ht]
\begin{center}
\renewcommand{\arraystretch}{1.2}
\begin{tabular}{c|c|c|c|c|c|c|c|c}
\multicolumn{3}{c}{Stage 1} \vline & \multicolumn{4}{c}{Stage 2} \vline & \multicolumn{2}{c}{val mAP@0.5}  \\
\hline
dataset & +M(A) & +M(K) & dataset & +M(A) & +M(K) & finetune & AVA & Kinetics \\
\hline \hline
A & $\surd$ & $\times$ & A+K & $\surd$ & $\surd$ & $\surd$ & 35.67 & 28.59 \\
A+K & $\times$ & $\times$ & A+K & $\surd$ & $\surd$ & $\times$ & 35.91 & 33.35 \\
A+K & $\surd$ & $\times$ & A+K & $\surd$ & $\surd$ & $\surd$ & \textbf{36.88} & \textbf{33.44} \\
\end{tabular}
\end{center}
\caption{\textbf{Comparison of different training strategies for memory bank on AVA-Kinetics.} All models are trained with SlowFast-152 backbone and T-only head. Results are tested with 3 scales and horizontal flips. "+M(A)" and "+M(K)" indicate maintaining online memory bank of AVA and Kinetics during the training stage. "A" indicates training with AVA only, and "A+K" indicates training jointly with AVA-Kinetics.}
\label{tab:mem}
\end{table*}
\subsection{Main Results}
Table \ref{tab:main} shows our main results on AVA-Kinetics. We report results of models using different backbones, pretrained datasets, input formats relation heads and memory banks. Our best single model achieves 37.95 mAP and 35.26 mAP on AVA and Kinetics, respectively, and 38.43 mAP on AVA-Kinetics Crossover. For backbones, ir-CSN-152 achieves better performance compared with SlowFast-152, which may attribute to that ir-CSN-152 is pretrained with smaller downscale of $1/16$, while SlowFast-152 is $1/32$. For the pretrained dataset, pretraining on Kinetics700 dataset generally improves 1 mAP compared with those pretrained on Kinetics400. We also investigate results with different temporal resolutions, and it is shown that increasing the temporal resolution $\times 2$ could improve around 0.6 mAP. For S-only and T-only relation heads, they both achieve great improvement compared with linear head and similar results to each other. T-only head with memory bank increase mAP of about 0.8 mAP. Among those models, we select 15 models and ensemble their results with average voting. The ensembled result achieves 40.97 mAP on the validation set of AVA-Kinetics, and 40.67 mAP on the test set. It is noted that even without training on the validation set, we only drop 0.3 mAP compared with the validation set.

\subsection{Ablation Studies and Discussions}

\textbf{Influence of memory bank strategies.} We train the three different strategies of maintain memory bank on AVA-Kinetics. As shown in Table \ref{tab:mem}, only training on AVA in the first stage and finetuning on AVA-Kinetics could achieve satisfying performance on AVA,  but the results on Kinetics will be poor since the backbones are not fully trained on this dataset. Training head without memory bank on both AVA and Kinetics in the first stage could improve the result on Kinetics in a large margin. However, this strategy must re-initialize the weights of head in the second stage and may lose some valuable training efforts. The final strategy trains on AVA-Kinetics on two stages and maintain empty bank in the first stage, which achieve the best performance and and used in our final version.

\begin{table}[]
\begin{center}
\renewcommand{\arraystretch}{1.2}
\begin{tabular}{r|l}
classes & mAP@0.5 diff. \\
\hline
\hline
top-20& - 0.16 \\
bottom-20& + 1.05 \\
all-60 & + 0.14\\
\hline
cut & + 22.57 \\
write & + 4.93 \\
text on cellphone & + 3.66 \\
\hline
get up & - 3.93\\
give/serve to & - 4.12 \\
work on computer & - 8.85 \\
\end{tabular}
\end{center}
\caption{\textbf{Influence of applying decoupled training for lone-tail learning.} Results of ir-CSN-152+T-only on AVA dataset are used for comparison. Classes are ranked by their numbers of labeled samples. Averaged results of the top-20 and bottom-20 classes, top-3 improved classes and bottom-3 improved classes are listed.}
\label{tab:decouple}
\end{table}

\textbf{Influence of decoupled learning.} We compare the results of all classes before and after the decoupled class-balanced finetuning. The classes are ranked by their number of samples, and we list the averaged difference of the top-20 and bottom-20 classes. As shown in Table \ref{tab:decouple}, there is an improvement of 1.05 mAP of those small classes. At the same time, the performance of the top-20 classes almost do not drop. We also report classes with largest changes after decoupled training. It is noted that some of the small classes, \eg cut, could get great improvement, while there also exists classes with dropped performance. This may attribute to the overfitting of those classes with increased training times, which could be further improved.

\begin{table}[]
\begin{center}
\renewcommand{\arraystretch}{1.2}
\begin{tabular}{c|c|c|c|c}
\multirow{2}{*}{detector} & \multicolumn{2}{c}{AVA} \vline & \multicolumn{2}{c}{Kinetics} \\
\cline{2-5}
& AP@0.5 & mAP@0.5 & AP@0.5 & mAP@0.5 \\
\hline \hline
det-v1 & 94.5 & 35.72 & 77.0 & 32.70 \\
det-v2 & 95.1 & 36.16 & 81.7 & 33.84\\
det-v3 & 95.1* & 36.41 & 82.3 & 34.80 \\
\hline
GT & - & 43.53 & - & 48.49\\
\end{tabular}
\end{center}
\caption{\textbf{Influence of person detector on AVA and Kinetics validation sets.} The column of AP@0.5 shows the detection results of different detectors. Results using GT bounding boxes are listed for reference. All the results are trained with ir-CSN-152 backbone and T-only head, and are tested with 3 scales and horizontal flips. "*" notes the results of det-v3, which has similar AP@0.5 to det-v2 but achieves better performance when measuring AP@0.5:0.95.}
\label{tab:box}
\end{table}
\textbf{Influence of human detector.} Here we have also investigate how much the improvement of the detector performance will gain the final action detection performance. Three detection boxes with 77.0\%, 81.7\%, 82.3\% AP on Kinetics subset of AVA-Kinetics, respectively, are applied in our pipeline. As Table~\ref{tab:box} shown, when the performance of the detector increases from 77\% to 81.7\%, the final performance can obtain significant improvement of about 1 mAP on both AVA and Kinetics parts of the dataset. If the GT boxes are applied, the performance of action detector will reach 43.5 mAP and 48.5 mAP respectively on the two subset. Moreover, even the improvements of human detector are similar (95.1 \textit{v.s.} 95.1 and 81.7 \textit{v.s.} 82.3 on AVA and Kinetics subset respectively), the final action detector also obtains corresponding gains of performance. The results demonstrate that human detector is one of bottleneck modules in the action detection pipeline.

\section*{Acknowledgement}
This work was supported by Alibaba Group through Alibaba Innovative Research Program.

{\small
\bibliographystyle{ieee_fullname}
\bibliography{egbib}
}

\end{document}